\pdfoutput=1
\documentclass[10pt,twocolumn,letterpaper]{article}
\usepackage{cvpr}

\usepackage[pagebackref=true,breaklinks=true,letterpaper=true,bookmarks=false]{hyperref}
\cvprfinalcopy 
\def\cvprPaperID{1893} 

\usepackage{amsmath, amsthm, amsfonts, amssymb}
\usepackage{graphicx}
\usepackage{wrapfig}
\usepackage{subfig}
\usepackage{color}
\usepackage{xspace}
\usepackage{subfig}
\usepackage{wrapfig}
\usepackage{enumitem}
\usepackage{dblfloatfix}
\usepackage{titling}

\def\shortcite{\cite}

\definecolor{turquoise}{cmyk}{0.65,0,0.1,0.1}
\definecolor{purple}{rgb}{0.65,0,0.65}
\definecolor{dark_green}{rgb}{0, 0.5, 0}
\definecolor{orange}{rgb}{0.8, 0.6, 0.2}
\definecolor{red}{rgb}{0.8, 0.2, 0.2}
\definecolor{brown}{rgb}{0.5, 0.16, 0.16}

\def\wunet{\mbox{WU-Net}\@\xspace}

\makeatletter
\renewcommand{\paragraph}{%
  \@startsection{paragraph}{4}%
  {\z@}{1.25ex \@plus .2ex \@minus .2ex}{-1em}%
  {\normalfont\normalsize\bfseries}%
}
\makeatother
\setlist{topsep=2mm, itemsep=0.5mm, parsep=0mm}

\begin{document}

\title{{\bf \Large Tags2Parts: Discovering Semantic Regions from Shape Tags} \vspace{-1mm}}
\date{\vspace{-4mm}}

\author{
  Sanjeev Muralikrishnan$^1$
\and
  Vladimir G. Kim$^2$
\and
  Siddhartha Chaudhuri$^{1, 2}$
\and
  \\
  $^1$IIT Bombay \,\,\,\,\,\, $^2$Adobe Research
}

\maketitle

\begin{figure*}[h!]
\input{teaser}
\end{figure*}

\begin{abstract}

\vspace{-2mm}

We propose a novel method for discovering shape regions that strongly correlate with user-prescribed tags. For example, given a collection of chairs tagged as either ``has armrest'' or ``lacks armrest'', our system correctly highlights the armrest regions as the main distinctive parts between the two chair types.
To obtain point-wise predictions from shape-wise tags we develop a novel neural network architecture that is trained with tag classification loss, but is designed to rely on segmentation to predict the tag. Our network is inspired by U-Net, but we replicate shallow U structures several times with new skip connections and pooling layers, and call the resulting architecture \emph{\wunet}.
We test our method on segmentation benchmarks and show that even with weak supervision of whole shape tags, our method can infer meaningful semantic regions, without ever observing shape segmentations. Further, once trained, the model can process shapes for which the tag is entirely unknown. As a bonus, our architecture is directly operational under full supervision and performs strongly on standard benchmarks. We validate our method through experiments with many variant architectures and prior baselines, and demonstrate several applications.

\vspace{-4.5mm}

\end{abstract}


\section{Introduction}
\label{sec:intro}

\vspace{-0.5mm}

Online repositories contain millions of 3D shapes, providing rich data for data-driven 3D analysis and synthesis~\cite{kai2015star}. While these repositories often provide tags, textual descriptions, and soft categorization to facilitate text-based search, these labels are typically provided for the entire shape, and not at the region level.
Many applications require finer shape understanding, e.g. parts and their labels are essential for assembly-based modeling interfaces.
While one can obtain these labels by training a strongly supervised segmentation model~\cite{Kalogerakis17}, this level of supervision requires substantially more involved annotation interfaces and human effort, making it infeasible for massive and growing online repositories.
Existing methods for discovering semantic regions without explicit supervision are typically guided by geometric cues (e.g.~\cite{Huang2011}), but they are prone to failure by being tailored to specific notions of parts, implicitly encoded by algorithm design.

{\em Weakly-} or semi-supervised methods have been proposed as a compromise between supervised and unsupervised techniques. For example, Yi et al.~\shortcite{Yi17} leverage scene graph metadata in existing repositories, which provide some segments and labels for a small subset of shapes. This metadata is very sparse and specific to computer graphics models. In contrast, tags for entire shapes are abundant, often accompany scanned shapes, and are easy to crowdsource at scale.


In this work we propose a novel method for discovering regions from shape tags without explicit region-wise labeling or prior segmentation. For example, in a collection of shapes tagged as ``has armrest'' and ``does not have armrest'', we are able to identify the armrest components of the chairs in the former category (Figure \ref{fig:teaser}). Further, once trained, our method can process shapes for which the tag is entirely unknown.
%
%

Our main challenge is that the weak supervisory signal (whole object tag) is different from the target output (point-wise labels). To address this, we use a neural network that jointly performs classification and segmentation, and train it for whole object tagging while relying on point-wise labels to infer the tags.

\begin{figure*}[t!]
  \includegraphics[width=\linewidth]{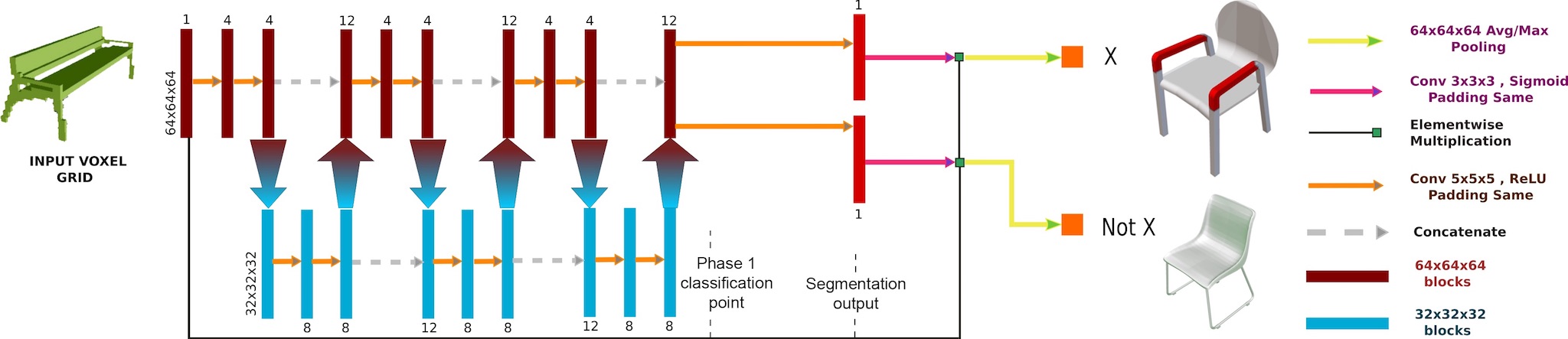}
  \caption{\wunet architecture, showing three stacked down/upsampling `U' structures linked by skip connections, ending in segmentation branches. Under weak supervision, the network is trained with only a classification loss.}
  \label{fig:arch}
  \vspace{-2mm}
\end{figure*}

In particular, we propose a novel neural network architecture with skip connections, which we call \mbox{\emph{\wunet}} (Figure~\ref{fig:arch}), inspired by the U-Net~\cite{Ronneberger15} architecture for {\em strongly}-supervised image segmentation. We make two key modifications. {\em First}, to regularize the network and improve localization of segments we replicate the `U' structure thrice (`WU') and add skip connections both within {\em and} across them. {\em Second}, since the network is originally designed for strongly-supervised segmentation, we add two layers for tag classification from a hidden segmentation layer: average pooling followed by max pooling. Average pooling encourages coherence, forcing the network to train for segments that help tag classification. This network architecture is our main technical contribution.

To evaluate our approach, we use shapes from standard datasets~\cite{Yi16}, but withhold region labels and only tag shapes based on part presence and absence. Our method detects regions with remarkable accuracy without observing a single segmented shape. As a bonus, we observe that our approach is also suitable for strongly-supervised segmentation, and demonstrate that the architecture performs well under strong supervision. We validate our design through extensive experiments, including ablation studies, variant architectures, baseline comparisons, and prototype applications.

\section{Related work}
\label{sec:related}
We overview related work on shape and image segmentation with various degrees of supervision.

\paragraph{Unsupervised shape segmentation}
One can leverage shape similarities and geometric cues to discover parts \cite{Golovinskiy09,Sidi2011,Huang2011}. These methods encode generic part priors, but not all semantic regions conform to them.
%
%
To bias unsupervised methods towards semantic regions, Yi et al.~\shortcite{Yi17} leverage existing scene graphs, which are sparse and not always informative.
%
No such method provides significant output control, which prevents discovery of user-prescribed regions.

\paragraph{Supervised shape segmentation}
A direct remedy is to use shapes with manually-labeled regions to train a model that can discover similar semantic regions in new shapes~\cite{Kalogerakis2010}. Recent methods exploit deep neural networks, based on 2D renderings~\cite{Kalogerakis17}, local descriptors after spectral alignment~\cite{yi2016syncspeccnn}, unordered point sets~\cite{qi2016pointnet}, canonicalized meshes~\cite{Maron17}, and voxel octrees~\cite{Riegler17,Wang17}.


The need to collect labeled data is the main bottleneck for supervised methods. Several approaches try to minimize this cost. E.g., Wang et al.~\shortcite{Wang2012} actively choose the next shape to label that most benefits a supervised method. Yi et al.~\shortcite{Yi2016} partially replace annotation with (quick) verification. These techniques still require tedious manual segmentation. Our goal is to avoid this altogether with a less taxing form of supervision, known as \emph{weakly supervised analysis}.

\paragraph{Distinctive regions in shapes}
In prior work most relevant to ours, Shilane and Funkhouser~\shortcite{Shilane2007} use hand-crafted local descriptors to highlight regions common to a category and different across categories. We learn such a representation via a neural network directly from a voxelized shape, and apply it to fine-grained shape segmentation within a single category. Further, unlike \cite{Shilane2007}, our method directly applies to test shapes with unknown tags, since it implicitly performs classification. In evaluations, our method is significantly more accurate. In recent work, Hu et al.~\shortcite{Hu17} identify small local elements correlated with object styles.

\paragraph{Weakly-supervised image segmentation}
Several computer vision methods can localize object data from whole-image tags (e.g.~\cite{Song14,Wang2014,Cinbis15}).
%
With the rise of deep neural networks, researchers observed that neurons in a classification network often activate on salient objects~\cite{Simonyan13}. Oquab et al.~\shortcite{Oquab15} append global max-pooling to a convolutional segmentation network~\cite{Long15} to obtain a classification network suitable for object localization.
%
In our work we focus on segmentation, and found that global max-pooling does not favor detecting coherent regions: we prefix it with average pooling for smoother segmentation. We also found that \wunet's skip connections improve results over sequentially stacked convolutions.
Pathak et al.~\shortcite{pathak15} study additional constraints, which we can potentially incorporate.


\section{Method}
\label{sec:method}

\subsection{Data representation}
We represent a 3D shape in voxelized form. Given a $64{\times}64{\times}64$ cubical grid tightly fitting the shape, we set each voxel intersecting the surface to $1$, and the rest to $0$. We omitted interior voxels. Apart from being a natural domain for 3D convolution, this representation ensures we do not take advantage of inherent part structure in meshes. In fact, our input need not be a mesh at all, as long as we can densely sample it.

\subsection{Network architecture}
Our method for weakly-supervised 3D shape segmentation utilizes a novel feedforward neural network architecture, which we call \wunet. It is inspired by the U-Net architecture of Ronneberger et al.~\shortcite{Ronneberger15}, which was proposed as an effective way to segment biomedical images with limited training data in a strongly supervised setting. U-Net's prominent feature, from which it derives its name, is a sequence of fully convolutional downsampling layers (the ``contracting'' arm of an `U'), followed by an inverse sequence of fully convolutional upsampling layers (the ``expanding'' arm of the `U'), with the two sequences bridged by skip connections.

The \wunet architecture leverages this building block by linking three fully convolutional U structures in sequence, i.e. a `W' followed by an U (Figure \ref{fig:arch}). Data flowing through the network therefore goes through three successive cycles of down- and up-sampling, from \mbox{$64^3$} to \mbox{$32^3$} and back to \mbox{$64^3$}, encouraging spatial coherence and spread in the detected signal. Unlike U-Net, our U's are very shallow, each one involving a single downsampling/upsampling sequence. We explain this design choice below.

Our architecture also has skip connections like U-Net, which allow reasoning in later layers to be sensitive to structure in the original data which may have been lost during downsampling. {\em Unlike} the original U-Net, the \wunet skip connections also provide bridging connections between {\em different} U structures (see the dashed arrows in the lower row of layers in Figure \ref{fig:arch}). This provides an elegant symmetry between the high and low resolution paths in the structure, with data winding back and forth between the resolutions while also having a secondary flow within the layers at each resolution. We also discuss this design choice below.

To map from a layer at one resolution to a layer at the same resolution (orange arrows), we employ several $5^3$ convolutional kernels. To downsample, we use a max pooling operator over a $2^3$ neighborhood. To upsample, we use bilinear interpolation of the feature map. All neurons in the `WU' structure have ReLU activations.

\paragraph{Output segmentation map.} The output of the final U is fed to two or more segmentation branches, one for each class. For weakly-supervised binary classification, e.g. ``has back'' vs ``lacks back'' for chairs, there are two branches. Under strong supervision, there is one branch for each part label: `seat', `back', `arm' etc. Each branch has one $3^3$ convolutional layer, with sigmoid activation. This layer acts as the {\em segmentation map} -- it is in one-to-one correspondence with the input, and its output values are interpreted as the probabilities of voxels having particular class labels.

\paragraph{Loss function.} Under strong supervision, a per-voxel cross-entropy loss is applied to the output segmentation map. Under weak supervision, we apply $2^3$ average pooling to this output, and then take the maximum over the pooled response. Average pooling encourages a wider response region (Section \ref{sec:results} tests the effect of other pooling radii). The max-pooled prediction (across branches) is compared to the GT shape tag with a cross-entropy loss.
To prevent activating empty voxels near shape boundaries, we multiply each segmentation map element by the corresponding element of the input voxel grid, letting the network focus only on errors over the shape.

\begin{figure}[b!]
  \centering
  \includegraphics[width=1.0\linewidth]{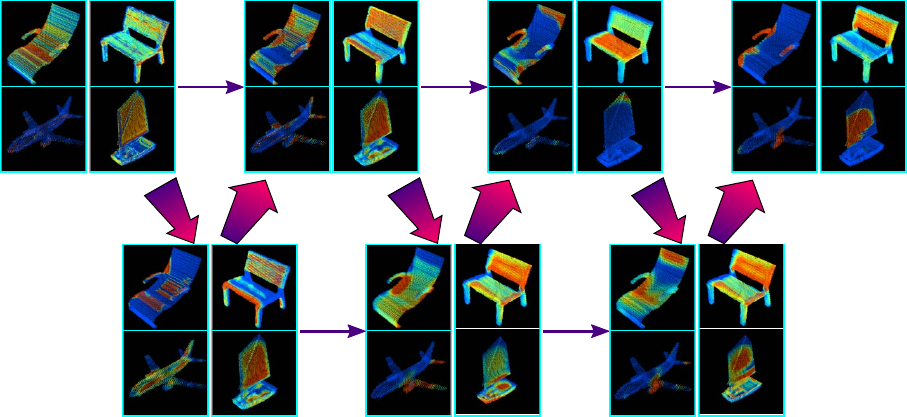}
  \caption{Information flow in weakly-supervised \wunet, captured as layer activation maps (red: high) for detecting chair arms/backs, ship sails, and plane engines. Shallow U's prevent over-summarization, and the signal is not lost by repeated concatenations from distant layers, unlike a deep U (Figure \ref{fig:deep_vis}).}
  \label{fig:wu_vis}
  \vspace{-4mm}
\end{figure}

\paragraph{Symmetrization.} Our dataset shows prominent symmetries, chiefly reflectional. Since such shapes have redundant local information, a classifier can achieve high accuracy without seeing the complete shape. \wunet is no exception, and our part detection often demonstrates consistent asymmetry,
\begingroup
\setlength{\intextsep}{3mm}
\setlength{\columnsep}{5mm}
\begin{wrapfigure}{r}{0.4\columnwidth}
  \vspace{-3mm}
  \includegraphics[width=\linewidth]{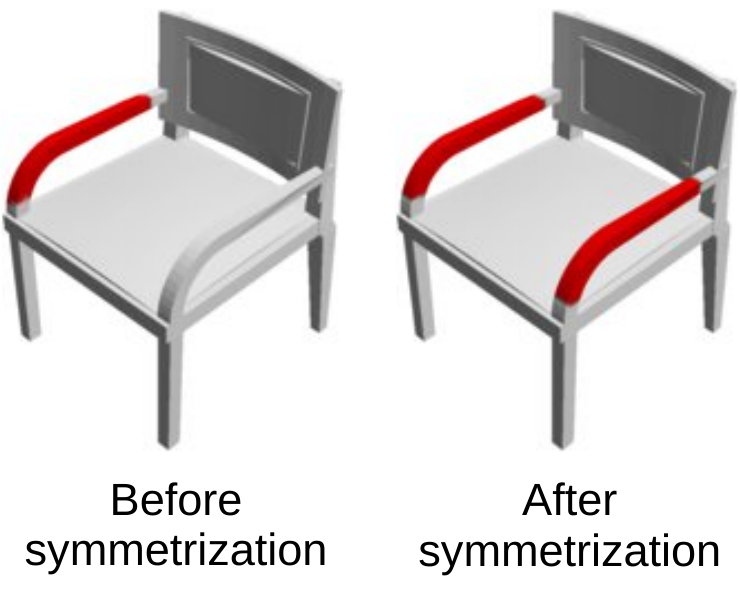}
  \label{fig:symmetry}
  \vspace{-6mm}
\end{wrapfigure}
yielding high precision but lower recall, e.g. when only right arms of chairs are detected (inset). To correct this, we simply mirror inferred salient regions on both sides of the symmetry plane.

\paragraph{Discussion of design choices.} \wunet has three shallow U's instead of a single deep one, bridged by skip connections at both high and low resolutions. These design choices enable convolutional filters in later layers to have a high effective field of view (by composition with filters from preceding layers) even on high resolution data. Each shallow U mildly summarizes the signal and then immediately analyses it jointly with the unsummarized signal. The information flow is visualized in Figure \ref{fig:wu_vis}. The ``low resolution'' skip connections provide each summarization step context from previous summaries. Section \ref{sec:results} shows that each successive stacked U improves performance, and weakly supervised shallow U's outperform deep U's.
\endgroup

Note the contrast to U-Net, where the latter half of the deeper architecture reconstructs successively higher resolution signals from a single drastic summary in the bottleneck layer. While skip connections do provide access to undecimated signals, the results of the joint high- and low-resolution analysis at each level are not further summarized, but simply upscaled to the next level. The filters in the final layer cannot have a high field of view on the original signal unmodified by downsampling. Thus, only excessively local information is incorporated from early layers by concatenation, which can drown out meaningful signals from the summarization layers when only weak supervision is available.
\begingroup
\setlength{\intextsep}{3mm}
\setlength{\columnsep}{5mm}

\begin{wrapfigure}{r}{0.55\columnwidth}
  \includegraphics[width=\linewidth]{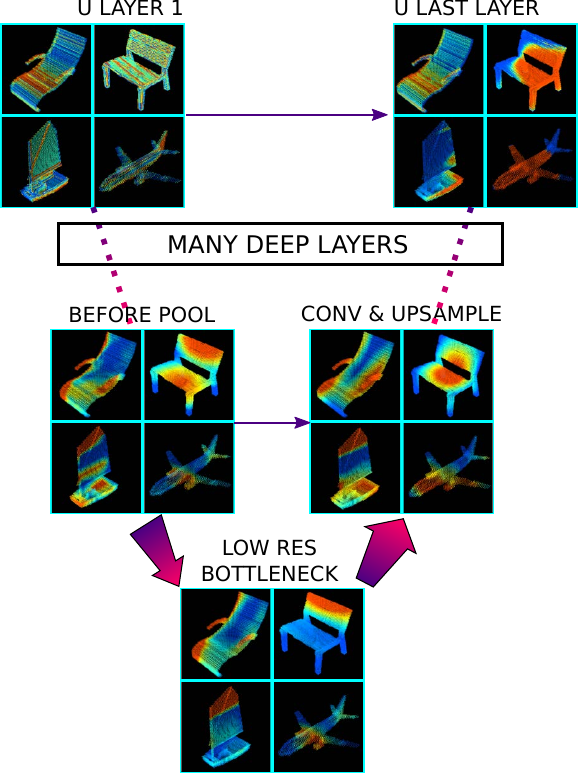}
  \vspace{-3mm}
  \caption{Arm, back, sail and engine signals are lost via over-summarization and ambiguous high-resolution concatenations in a weakly-supervised deep U-Net.}
  \label{fig:deep_vis}
  \vspace{-1mm}
\end{wrapfigure}
%

In Figure \ref{fig:deep_vis}, we show how activation maps in a single deep U suffer from excessive summarization, which the weak supervisory signal is not sufficient to repair despite skip connections: per-voxel strong supervision is required. Even though the bottleneck layer correctly localizes parts, multiple rounds of subsequent upsampling and ambiguous detail introduced from earlier layers spread the signal out incorrectly.

\subsection{Training}

In the weakly supervised setting, the \wunet architecture is trained in two phases. We found the two-phase training to give better results than a single phase alone. The phases are described below.
\endgroup

\paragraph{Phase 1 (no output segmentation map).} In this phase, the final segmentation branches are removed and a simple classification layer is temporarily appended to the `WU'. This layer computes the maximum, over all voxels, of each of the 12 `WU' output channels, followed by a fully-connected map from the 12 maxima to two outputs (the complete shape label, e.g. ``armrest'' vs ``no armrest'').  This network is trained with cross-entropy loss until the classification accuracies on both training and validation sets exceed 95\%. Once this happens, we adjudge the network to have high generalization accuracy and move to the next phase. Further phase 1 training tends to overfit.

\begin{figure}[t!]
  \centering
  \includegraphics[width=0.8\linewidth]{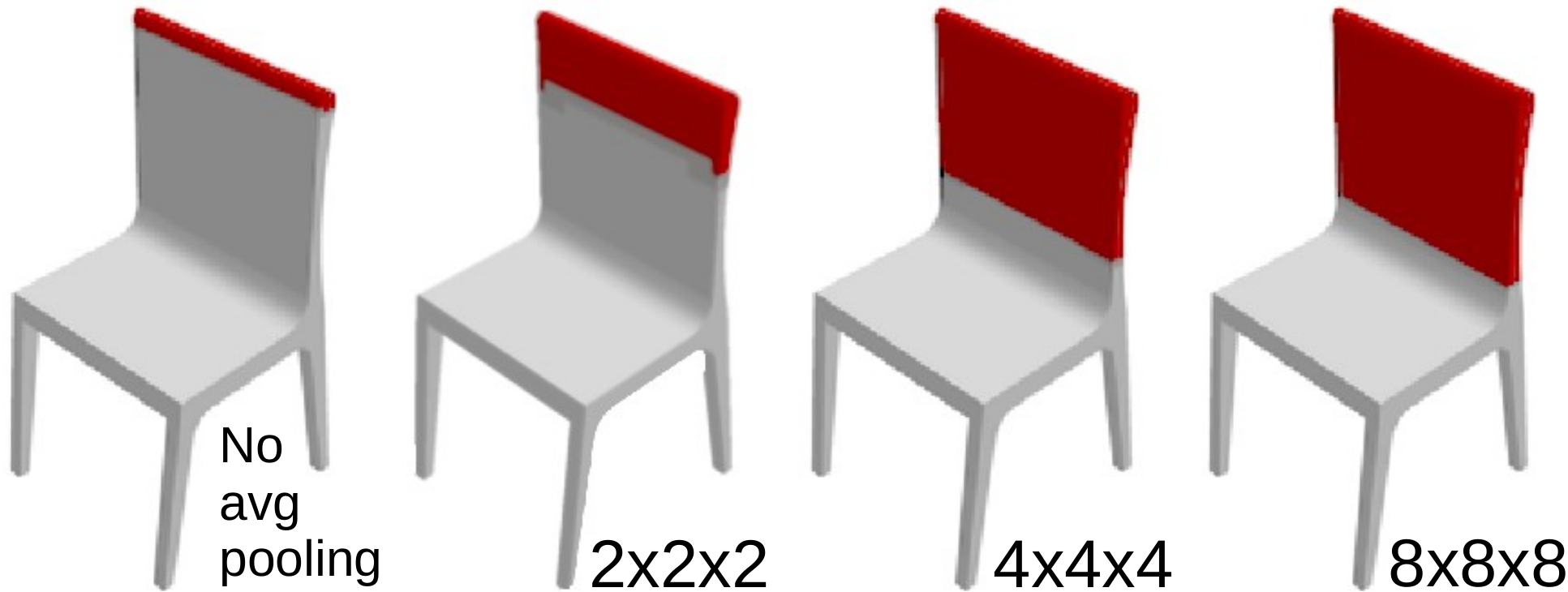}
  \caption{The effect of increasing the kernel size in average pooling. While here the largest kernel works best, for categories with finer parts this is not so.}
  \label{fig:pooling_vis}
  \vspace{-3mm}
\end{figure}

\paragraph{Phase 2 (with output segmentation maps).} We now remove the phase 1 classification layer, restore segmentation branches, and train the whole network end-to-end. We found benefit in slowly enlarging the average-pooling kernel, starting from 0 (no pooling) for 50 epochs, followed by 10 epochs for each expansion of the kernel. The best overall performance came from a $2^3$ final kernel, and we report all comparative results with this setting. However, for specific datasets larger kernels may help further, as we show in our evaluation. Generally, detection of larger salient parts is aided by larger average pooling kernels (Figure \ref{fig:pooling_vis}).

Under {\em strong} supervision, we dispense with two-phase training and final pooling, and directly train the network end-to-end with a per-voxel cross-entropy loss over the segmentation maps for each output label.

%
%
%
%

\section{Results}
\label{sec:results}

We evaluate our method on standard datasets that contain various semantic region labels. For weakly-supervised segmentation, which is our principal focus, our extensive comparisons suggest that \wunet is a ``sweet spot'' in the space of related architectures. We also show that the same network performs strongly under strong supervision on standard benchmarks.


\subsection{Weakly Supervised Region Labeling}
\label{sec:weaksup}

In these validation experiments, we test if our \wunet architecture successfully detects salient parts that distinguish one category of shapes from another. We collated six different pairs of fine-grained shape classes, each pair distinguished by a prominent semantic component. These classes were: (a) {\em chairs} with and without {\bf armrests}, (b) {\em chairs} w/wo {\bf backs}, (c) {\em cars} w/wo {\bf roofs}, (d) {\em airplanes} w/wo wing-mounted {\bf engines}/propellers, (e) {\em ships} w/wo {\bf sails}, and (f) {\em beds} w/wo {\bf heads}. Fine-grained classes (a-d) are available in ShapeNet~\cite{Chang2015} with ground-truth labeled segmentations, although we had to collect chairs without backs (stools) from ModelNet~\cite{Wu15}. We annotated ships (e) and beds (f) ourselves.

Each class was randomly split 50:50 into train/test sets. Under weak supervision, segmentation/labeling accuracy on the training set is as important as on the test set. Still, the test set allows us to directly compare with a strongly supervised baseline. Dataset statistics are in Table \ref{tab:weakdata}. The meshes were voxelized with Binvox~\cite{Binvox}. Our data (and code) will be publicly available.

In evaluations presented in this section we report {\bf area under the curve (AUC)} for precision/recall curves,
where {\bf higher} numbers indicate {\bf better} performance. Our method is labeled ``\wunet + symmetrization''.
Full plots are in supplementary material.

\begin{table}[t!]
\small
 \centering
    \begin{tabular}{@{}c@{}|@{}c@{}|@{}c@{}|@{}c@{}}
      \,\,Shape category\,\, & \,\,Part category\,\, & \,\,Has part\,\, & \,\,Lacks part\,\, \\
      \hline
      \hline
      (a) Chair & Armrest & 481 & 1359 \\
      \hline
      (b) Chair & Back & 150 & 75 \\
      \hline
      (c) Car & Roof & 806 & 106 \\
      \hline
      (d) Airplane & Engine & 1034 & 266 \\
      \hline
      (e) Ship & Sail & 95 & 674 \\
      \hline
      (f) Bed & Head & 19 & 4 \\
      \hline
    \end{tabular}
    \vspace{-2.5mm}
  \caption{Weakly supervised segmentation dataset.}
  \label{tab:weakdata}
  \vspace{-5.5mm}
\end{table}

\paragraph{Segmentation performance.} Table \ref{tab:ablation} reports the per-voxel labeling accuracy of \wunet with identical hyperparameters (including $2^3$ avg pooling and symmetrization) and automatic training protocol in the 6 weakly supervised segmentation tasks, on training shapes. (Training set segmentation accuracy is a relevant performance metric under weak supervision. When computing it, we do not use ground truth tags. Test set performance is similar, see supplementary.)
For comparison we use these ablated alternatives:
\vspace{-0.8mm}
\begin{itemize}
  \item The {\em saliency map} of the trained \wunet, computed as the gradient of output w.r.t. input.
  \item \wunet\ {\em without skip connections}, representing a conventional fully convolutional architecture.
  \item \wunet\ {\em without the final U}, dubbed W-Net.
  \item \wunet\ {\em without 2 of the U's}, just a single shallow U, dubbed V-Net.
\end{itemize}
\vspace{-0.8mm}
Further, we also test \wunet without symmetrization.

\wunet, with or without symmetrization, substantially improves upon these alternatives. Training of the ablated networks does not always converge. When it does converge, V-Net and W-Net perform reasonably well, though they don't match \wunet. The version without skip connections is much worse.

In Table \ref{tab:deepu}, we compare \wunet with networks using deep U's. All outputs are symmetrized.
\vspace{-0.8mm}
\begin{itemize}
  \item A 3D analogue of the original U-Net~\cite{Ronneberger15}, with {\em one deep U structure} that repeatedly halves the grid resolution to $4^3$, then repeatedly doubles it back to $64^3$, with skip connections at every resolution.
  \item {\em 2 and 3 deep U's}, linked with high and low skip connections just like \wunet.
	\item The above 3 networks, with {\em Inception-style blocks}~\cite{Szegedy15} at every layer. Each $5^3$ kernel has $2^3$ and $3^3$ kernels also applied in parallel.
  \item A 3D version of a {\em Stacked Hourglass Network} (SHN$_\text{3D}$)~\cite{Newell16}, modeled as 3 deep U's without low-resolution skip connections between different U's.
\end{itemize}
\vspace{-0.8mm}
Deep 3D U-Net training converges, but it identifies incorrect parts (e.g. chair seats, not backs). Apart from a single deep U for chair armrests and a double deep U for backs, the rest cannot identify meaningful parts. This validates our use of shallow U's for weakly supervised segmentation.

\begin{figure*}[!b]
  \vspace{-2mm}
  \includegraphics[width=1.0\linewidth]{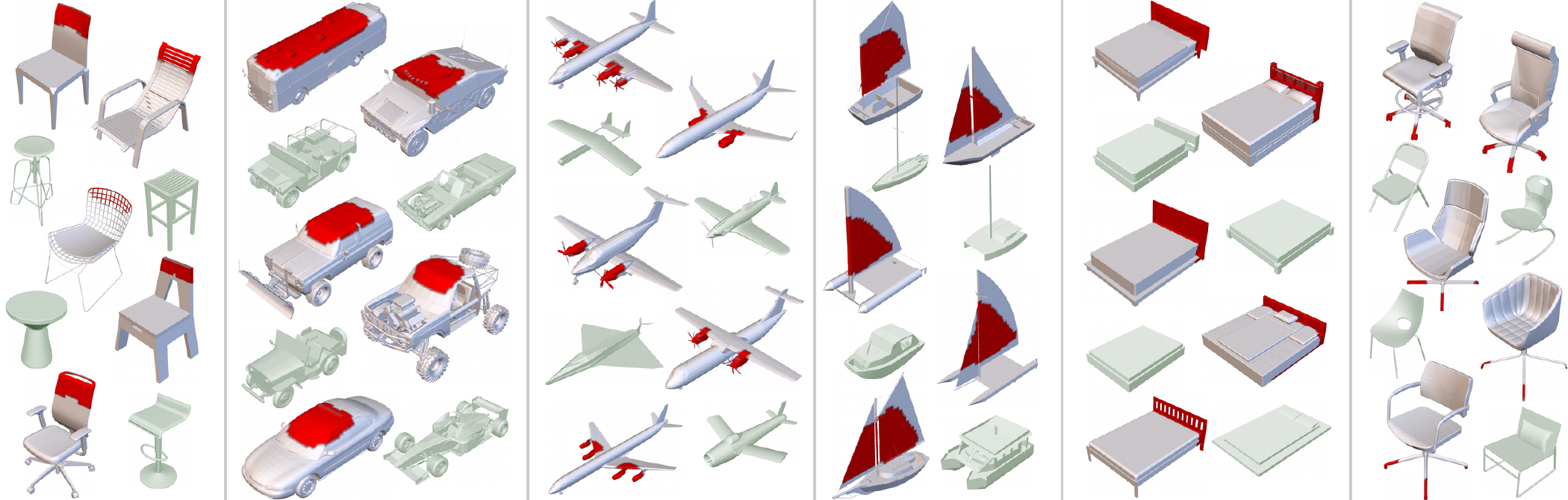}
  \caption{Examples of weakly supervised segmentation by \wunet. Left to right: detecting chair backs, car roofs, wing-mounted airplane engines/propellers, ship sails, bed heads, and chair swivels.}
  \label{fig:mixed}
  \vspace{-5mm}
\end{figure*}

We also present visual examples of symmetrized \wunet output, for a threshold of 0.9, in Figures \ref{fig:teaser} and \ref{fig:mixed}. In addition we also show some visual results on swivel chairs, for which ground truth segmentations were not available: the roller wheels were identified as salient in these shapes. Visual results on all shapes in our datasets are provided in supplementary material.

\begin{table}[!t]
\small
 \centering
    \begin{tabular}{@{}c@{}||@{}c@{}|@{}c@{}|@{}c@{}|@{}c@{}|@{}c@{}|@{}c@{}}
      & ~Arm~ & ~Back~
      & ~Roof~ & ~Propeller~ & ~Sail~
      & ~Bed~ \\
			\hline
			\hline
			WU-Net\, & \textbf{0.69} & \textbf{0.79} & 0.32 & \textbf{0.46} & \textbf{0.77} & \textbf{0.32} \\
			+ symmetrization\, & & & & & & \\
			\hline
			WU-Net\,  & 0.61 & 0.76 & \textbf{0.39} & 0.39 & 0.76 & \textbf{0.32} \\
			\hline
			W-Net\,  & 0.54 & 0.73 & 0.09 & 0.06 & 0.55 & 0.15 \\
			\hline
			V-Net\,  & 0.60 & 0.76 & 0.03 & 0.34 & 0.52 & 0.12 \\
			\hline
			No Skip Connections\,  & 0.07 & 0.62 & 0.05 & 0.09 & 0.30 & 0.17 \\
			\hline
			Gradient saliency\, & 0.03 & 0.27 & 0.12 & 0.18 & 0.00 & 0.29 \\
			+ symmetrization\, & & & & & & \\
			\hline
    \end{tabular}
  \caption{AUC of WU-Net vs various ablations for weakly-supervised segmentation (on training shapes).}
  \label{tab:ablation}
  \vspace{-2mm}
\end{table}

\begin{table}[!t]
\small
 \centering
    \begin{tabular}{@{}c@{}||@{}c@{}|@{}c@{}|@{}c@{}|@{}c@{}|@{}c@{}|@{}c@{}}
      & ~Arm~ & ~Back~
      & ~Roof~ & ~Propeller~ & ~Sail~
      & ~Bed~ \\
			\hline
			\hline
			WU-Net\, & \textbf{0.69} & \textbf{0.79} & \textbf{0.32} & \textbf{0.46} & \textbf{0.77} & 0.32 \\
			+ symmetrization\, & & & & & & \\
			\hline
			3 Deep U\,  & 0.03 & 0.19 & 0.03 & 0.01 & 0.22 & 0.21 \\
			(Inception)\, & & & & & & \\
			\hline
			2 Deep U\,  & 0.00 & 0.08 & 0.05 & 0.00 & 0.27 & 0.39 \\
			(Inception)\, & & & & & & \\
			\hline
			1 Deep U\,  & 0.04 & 0.19 & 0.04 & 0.14 & 0.54 & 0.13 \\
			(Inception)\, & & & & & & \\
			\hline
			3 Deep U\,  & 0.08 & 0.10 & 0.03 & 0.00 & 0.00 & \textbf{0.42} \\
			\hline
			2 Deep U\,  & 0.05 & 0.47 & 0.03 & 0.00 & 0.27 & 0.10 \\
			\hline
			1 Deep U\,  & 0.31 & 0.01 & 0.03 & 0.11 & 0.16 & 0.06 \\
			\hline
      SHN$_\text{3D}$\,  & 0.35 & 0.39 & 0.04 & 0.16 & 0.45 & 0.16 \\
			\hline
    \end{tabular}
  \caption{AUC of WU-Net vs Deep U-Net variants (symmetrized) for weakly-supervised segmentation (on training shapes).}
  \label{tab:deepu}
  \vspace{-2mm}
\end{table}

\begin{table}[!t]
\small
 \centering
    \begin{tabular}{@{}c@{}||@{}c@{}|@{}c@{}|@{}c@{}|@{}c@{}|@{}c@{}|@{}c@{}}
      & ~Arm~ & ~Back~
      & ~Roof~ & ~Propeller~ & ~Sail~
      & ~Bed~ \\
			\hline
			\hline
			WU-Net\, & 0.71 & 0.73 & 0.35 & 0.42 & \textbf{0.84} & 0.37 \\
			+ symmetrization\, & & & & & & \\
			\hline
			Strong supervision\, & 0.06 & 0.43 & 0.70 & 0.57 & 0.07 & \textbf{0.48} \\
			without classifier\, & & & & & & \\
			\hline
			Strong supervision\, & \textbf{0.91} & \textbf{0.97} & \textbf{0.89} & \textbf{0.89} & 0.68 & \textbf{0.48} \\
			with classifier\, & & & & & & \\
			\hline
    \end{tabular}
  \caption{AUC of Weakly supervised WU-Net vs a strongly supervised baseline (on test shapes).}
  \label{tab:strong}
  \vspace{-4mm}
\end{table}

\begin{table}[!t]
\small
 \centering
    \begin{tabular}{@{}c@{}||@{}c@{}|@{}c@{}|@{}c@{}|@{}c@{}|@{}c@{}|@{}c@{}}
      & ~Arm~ & ~Back~
      & ~Roof~ & ~Propeller~ & ~Sail~
      & ~Bed~ \\
			\hline
			\hline
			WU-Net\, & \textbf{0.69} & \textbf{0.79} & 0.32 & \textbf{0.46} & 0.77 & 0.32 \\
			+ symmetrization\, & & & & & & \\
			\hline
			SF 0.25\, & 0.43 & 0.45 & \textbf{0.51} & 0.12 & \textbf{0.87} & 0.53 \\
			\hline
			SF 0.5\, & 0.52 & 0.59 & 0.27 & 0.13 & 0.67 & \textbf{0.62} \\
			\hline
			SF 1.0\,  & 0.49 & 0.37 & 0.39 & 0.18 & 0.65 & 0.34 \\
			\hline
			SF 2.0\,  & 0.39 & 0.42 & 0.42 & 0.29 & 0.61 & 0.34 \\
			\hline
    \end{tabular}
  \caption{AUC of WU-Net vs Shilane and Funkhouser (SF) [16] at different scales (on training shapes). Note that SF
requires knowledge of ground truth tags at test time, whereas our method does not use them.}
  \label{tab:shilane}
  \vspace{-2mm}
\end{table}

\begin{table}[!t]
\small
 \centering
    \begin{tabular}{@{}c@{}||@{}c@{}|@{}c@{}|@{}c@{}|@{}c@{}|@{}c@{}|@{}c@{}}
      & ~Arm~ & ~Back~
      & ~Roof~ & ~Propeller~ & ~Sail~
      & ~Bed~ \\
			\hline
			\hline
			2x2x2 (default)\, & \textbf{0.69} & \textbf{0.79} & \textbf{0.32} & 0.46 & \textbf{0.77} & 0.32 \\
			\hline
			No avg pooling\,  & 0.63 & 0.67 & \textbf{0.32} & \textbf{0.54} & 0.70 & 0.31 \\
			\hline
			4x4x4\,  & 0.49 & 0.42 & 0.05 & 0.26 & \textbf{0.77} & 0.32 \\
			\hline
			8x8x8\,  & 0.08 & 0.23 & 0.00 & 0.01 & 0.58 & \textbf{0.33} \\
			\hline
    \end{tabular}
  \caption{The statistical effect (AUC) of increasing the kernel size for average pooling at the end of WU-Net.}
  \label{tab:pooling}
  \vspace{-4mm}
\end{table}

\paragraph{Comparison to a strongly supervised baseline.} For further insight, we train \wunet with strong supervision, with a single segmentation branch which is thresholded for a precision-recall plot. (We cannot use the strongly supervised variant of Section \ref{sec:fullsup}, because it outputs the max over label branches per voxel, and has no tunable threshold.) This strongly supervised network is not a classifier, and hence can end up identifying a semantic part in a shape which lacks it. This leads to very poor accuracy (Table \ref{tab:strong}). If we aid it by using the trained weak network simply as a binary classification oracle (${\sim}99\%$ accurate), then it establishes a high baseline as expected. This indicates the very valuable role shape tags play in identifying semantic parts.

\begin{figure}[!b]
  \vspace{-5mm}
  \includegraphics[width=1.0\linewidth]{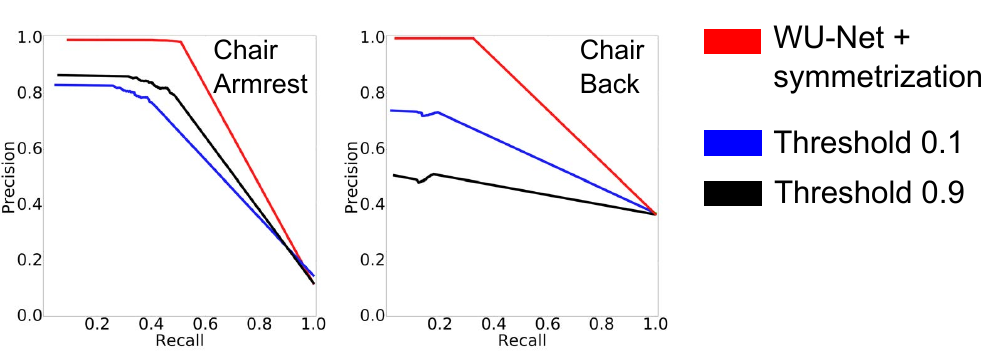}
  \caption{Chair armrest and back detected independently (red), vs detected in a multi-label setting for different thresholds on the label classifier outputs.}
  \label{fig:multilabel}
  \vspace{-5mm}
\end{figure}

\begin{table*}[!t]
\small
 \centering
    \begin{tabular}{@{}c@{}||@{}c@{}|@{}c@{}||@{}c@{}|@{}c@{}|@{}c@{}|@{}c@{}||@{}c@{}|@{}c@{}|@{}c@{}||@{}c@{}|@{}c@{}|@{}c@{}}
      & ~ShapeBoost~ & ~ShapePFCN~
      & ~1SU~ & ~2SU~ & ~{\bf \wunet} (3SU)~ & ~4SU~
      & ~1DU~ & ~2DU~ & ~3DU~
      & ~1DUI~ & ~2DUI~ & ~3DUI~ \\
      \hline
      \hline
      Co-aligned\,
      & 83.1 & 89.0
      & 87.8 & 89.8 & 90.2 & 90.0
      & 90.5 & 90.7 & 90.1
      & 91.2 & 90.9 & 91.3 \\
      \hline
      Randomly rotated\,
      & - & -
      & 73.1 & 74.7 & 75.1 & 75.6
      & 77.8 & 77.8 & 78.7
      & 78.3 & 79.0 & 79.0 \\
      \hline
    \end{tabular}
  \caption{Strongly-supervised segmentation and labeling accuracy (\%), averaged over 16 categories, for test shapes in ShapeNetCore, versus ShapeBoost~\cite{Kalogerakis2010} and ShapePFCN~\cite{Kalogerakis17}. Different {\wunet}-style variants have abbreviated names: 3SU is a sequence of {\bf 3} {\bf s}hallow {\bf U}'s (i.e. \wunet), 1DUI is {\bf 1} single {\bf d}eep {\bf U} ({\bf I}nception-style); each variant trained for 100 epochs. Full per-category statistics are in supplementary material.}
  \label{tab:fullsup}
  \vspace{-3mm}
\end{table*}

\paragraph{Comparison to Shilane and Funkhouser~\shortcite{Shilane2007}.} There is little prior work on weakly supervised 3D shape segmentation. The most relevant research is by Shilane and Funkhouser, who identify distinctive regions in different shape categories. While our scenario is slightly different (fine-grained intra-category differences), their method can be evaluated directly in our training setup. (Note that S-F relies on ground-truth shape tags to find distinctive regions. It does not work in our {\em test} setup, where the shape tag is unknown.) Table \ref{tab:shilane} shows results. The S-F results were not symmetrized -- symmetrization slightly hurt results because of false positives. In three out of six cases (chair arms/backs, airplanes), \wunet with default settings significantly outperforms S-F at all manually specified scale settings. In the other cases (cars, ships, beds), \wunet is a little worse, but the scales at which S-F outperform it turn out to be dramatically suboptimal in other cases. On average, \wunet significantly outperforms Shilane-Funkhouser at any given scale.

\paragraph{The role of the average-pooling layer.} The kernel size of the average-pooling layer after the segmentation map is a tunable hyperparameter that directly affects the identified regions in a visually interpretable way. For large semantic parts, a larger final kernel size often yields better results. The effect is one of degree, as seen in Table \ref{tab:pooling}, and depends on the data. However, we found that a fixed $2^3$ kernel achieves good performance in all cases, and this is the setting we present for our fully automatic method and for all evaluations.

\paragraph{Multi-label weak supervision.} What does \wunet predict when weakly supervised with {\em multiple} shape-level tags? While this is not this paper's focus, we find in preliminary investigations the framework can extend to some multi-label settings. For instance, we trained a single \wunet on chairs tagged with ``arm'' and ``back'', where either, both or no labels could be present. (In fact, we could not find chairs with arms but no backs.) This \wunet had two branches, one for each part label, and the segmentation map from each branch was output if the classification score exceeded a common threshold. In Figure \ref{fig:multilabel}, we show the multi-label output, vs training a different binary {\wunet} for each part. While the multi-label scores are competitive, especially considering the training data lacks ``arm but no back'' combinations, they do not exceed the binary results. Extending weakly supervised 3D segmentation to a range of multi-label scenarios is a ripe avenue for future work. A major difficulty is that in real datasets, weak tags are often strongly correlated (e.g. chair legs and seats). A small amount of strong supervision may resolve this.

\begin{figure*}[b!]
\centering
\begin{minipage}[t]{.28\linewidth}
  \centering
  \vspace{0pt}
  \includegraphics[width=1.0\linewidth]{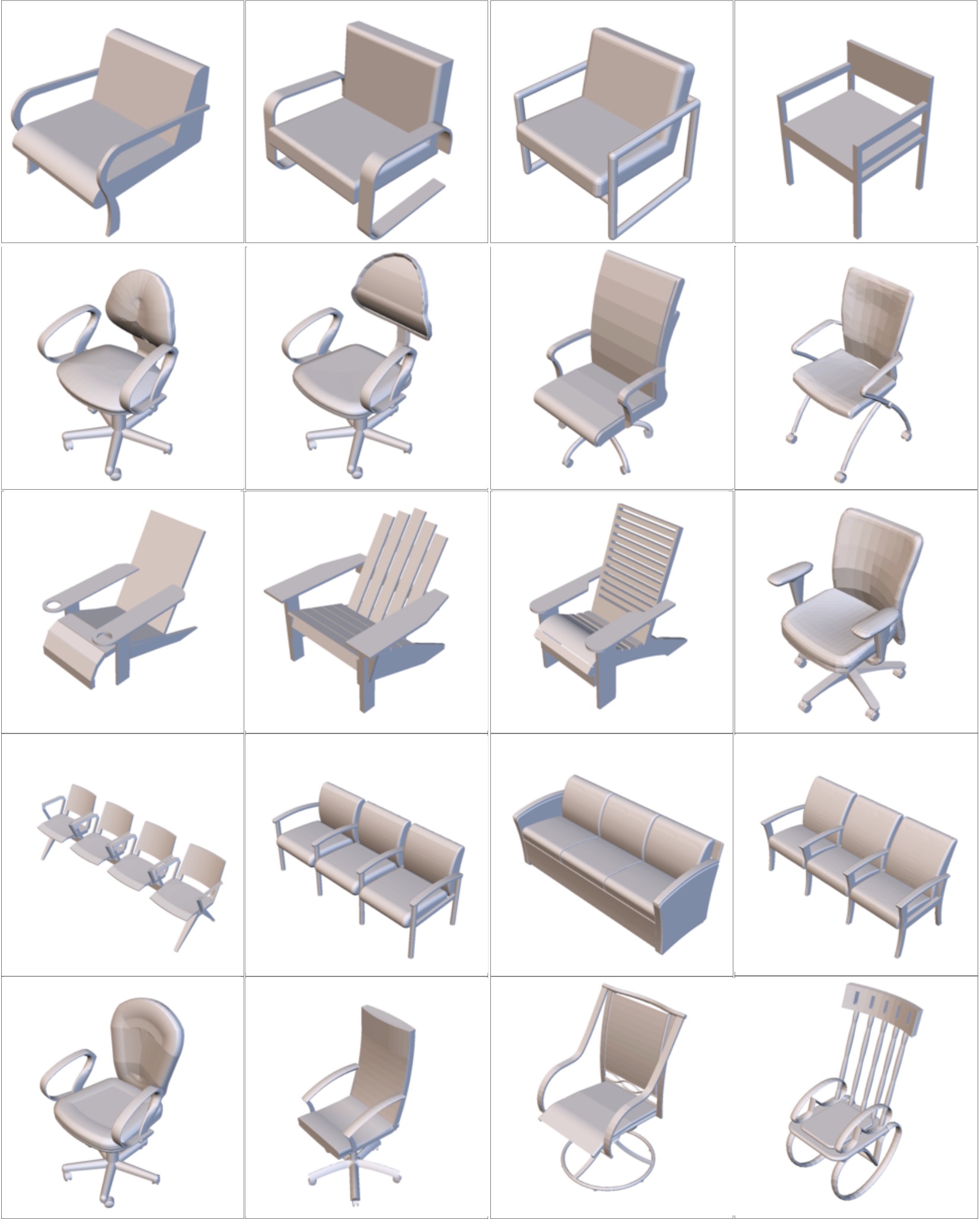}
  \captionof{figure}{Each row shows the top 3 shapes with similar armrests, detected by \wunet, retrieved for the query in the first column.}
  \label{fig:search}
\end{minipage}%
\hspace{.015\linewidth}
\begin{minipage}[t]{.38\linewidth}
  \centering
  \vspace{0pt}
  \includegraphics[width=1.0\linewidth]{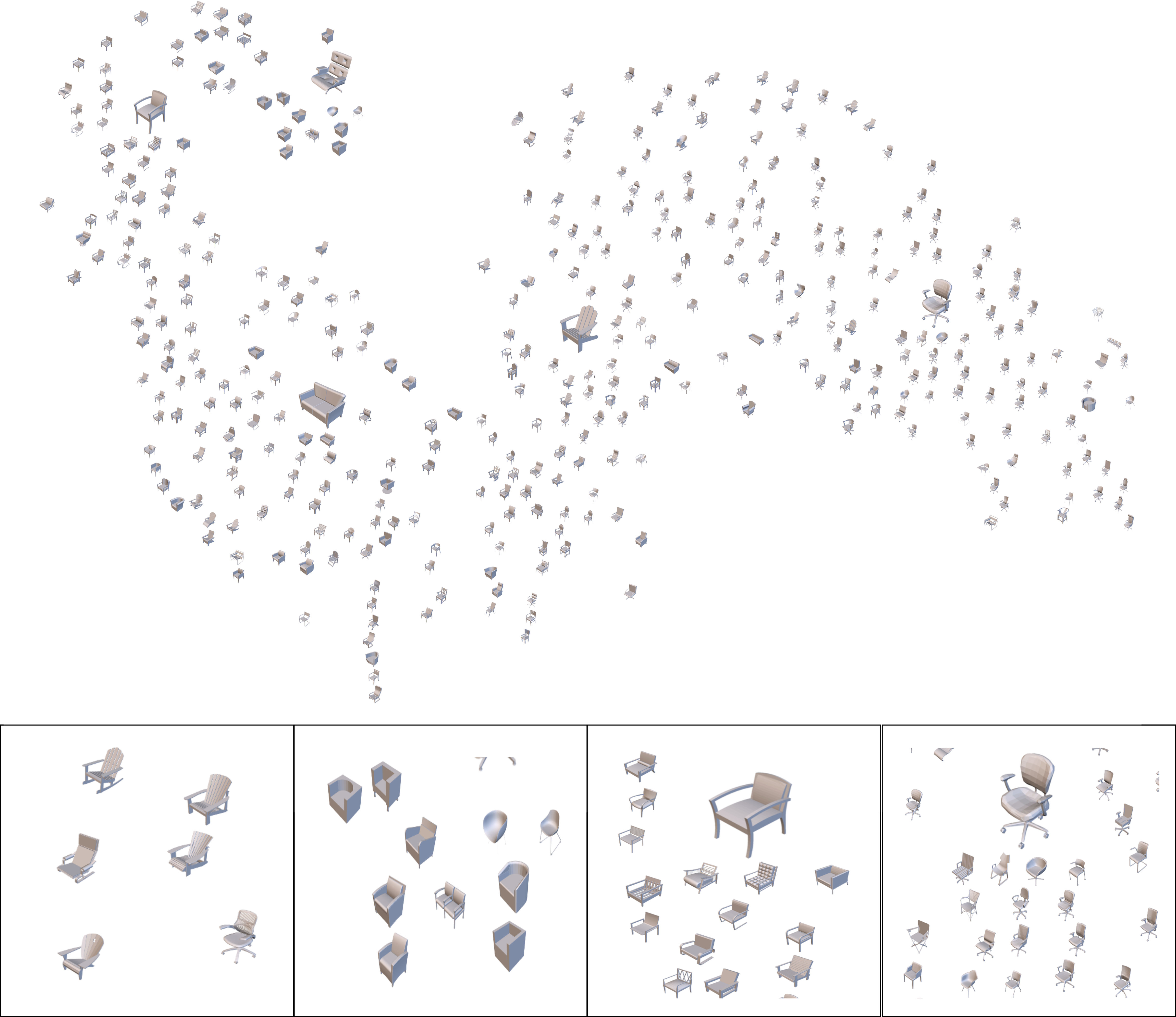}
  \captionof{figure}{A t-SNE embedding of chairs organized by similarity of the ``armrest'' regions detected by \wunet. Below, we show several zoomed-in regions of the image. The larger icons on top represent diverse representatives of the collection that can be obtained from this similarity metric.}
  \label{fig:embedding}
\end{minipage}
\hspace{.015\linewidth}
\begin{minipage}[t]{.28\linewidth}
  \centering
  \vspace{0pt}
  \includegraphics[width=1.0\linewidth]{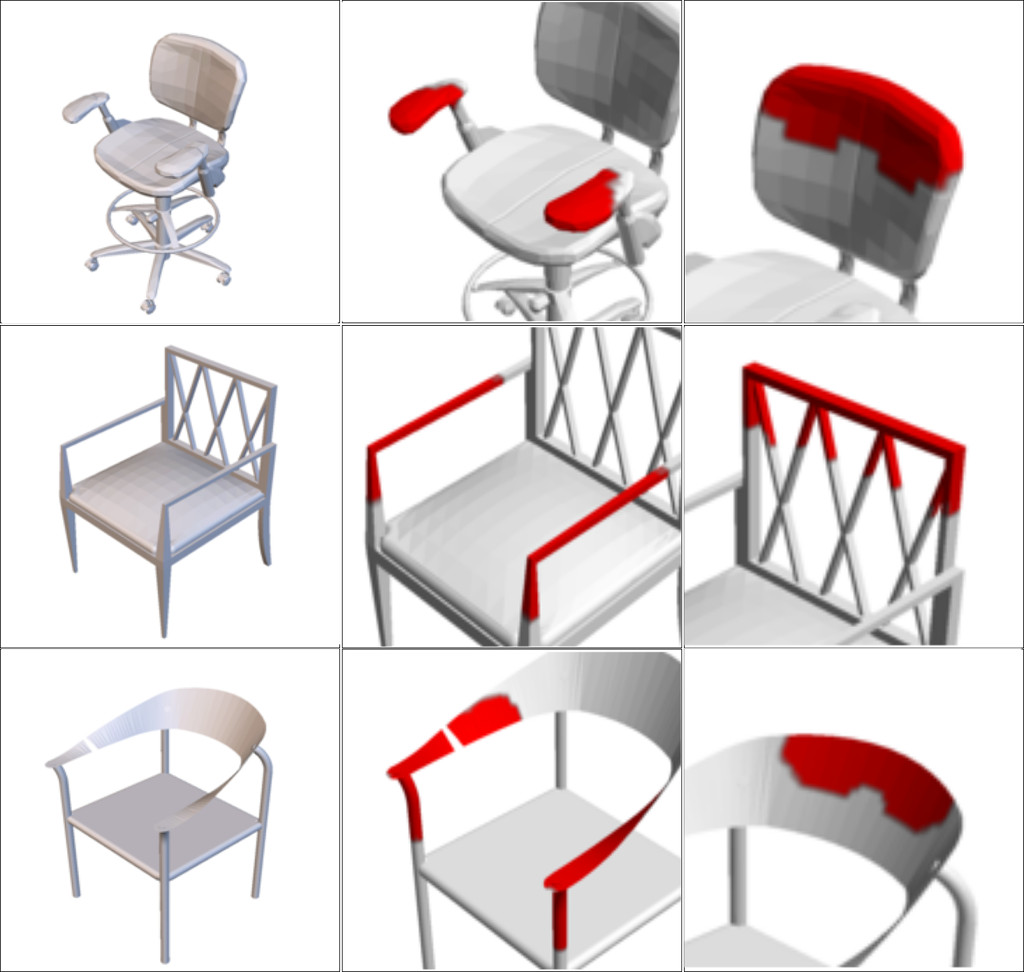}
  \captionof{figure}{Different thumbnails of the same shapes (first column) created to highlight detected ``armrest'' (second column) and ``back'' (third column) regions.}
  \label{fig:icon}
\end{minipage}
\vspace{-4mm}
\end{figure*}

\subsection{Strongly Supervised Region Labeling}
\label{sec:fullsup}

The \wunet architecture has the great advantage of being directly deployable in a strongly supervised setting, where per-point labels are available. We therefore test it on a standard benchmark: ShapeNetCore~\cite{Chang2015}. This dataset has manually annotated ground-truth segmentations for thousands of shapes in 16 categories. We compare our method to a recent state-of-the-art technique (\shortcite{Kalogerakis17}), using the same train/test splits. Our performance is summarized in Table \ref{tab:fullsup} (more details in supplementary material). Since the benchmark shapes are co-aligned, \wunet can take advantage of this to achieve state-of-the-art scores. However, \wunet is not designed to be rotation-invariant by default, unlike ShapePFCN~\cite{Kalogerakis17}. If we train and test with every shape independently and randomly rotated, scores drop by about \mbox{10-15\%}, since the network can no longer memorize rough absolute locations for parts. This is a familiar issue with voxel grid-based networks, and augmentation with many more rotations may help.

We also compare with a wide variety of variant architectures, described in the previous section, and present results in Table \ref{tab:fullsup} (and full per-category results in supplementary material). From these, we can infer the following (especially from results with rotation): {\em under strong supervision}, (a)~deep U's, (b)~stacking multiple U's, and (c)~Inception-style networks all improve performance a little. Combining all three factors yields the highest accuracy. Note that this improvement does not extend to weakly-supervised training.


\subsection{Applications}
\label{sec:apps}
We demonstrate the wide utility of our method with three mockup applications that focus on organizing a 3D shape database. First, we enable \emph{part-sensitive shape search} (Figure \ref{fig:search}) by computing a fine-grained shape similarity metric that focuses only on a user-selected tag (our simple implementation uses weighted average distance between the salient voxels, after aligning centroids of  salient regions). Since we map tags to specific geometric regions, we can make queries like: ``find chairs with similar armrests''. Second, we show \emph{fine-grained exploration of a shape dataset} (Figure \ref{fig:embedding}), demonstrating that the entire dataset can be organized based on similarity metrics computed for a specific tag (our prototype uses the simple metric above), providing users with different tag-focused views of the database. Third, we demonstrate that our method facilitates better~\emph{thumbnail creation} (Figure \ref{fig:icon}) by focusing on salient regions that correspond to specific tags. Automatically-generated thumbnails are commonly used for rapid browsing, and demonstrating important surface regions can provide better shape understanding for the user.


\section{Conclusion}
\label{sec:future}

We presented a method to obtain fine-grained semantic part annotations of 3D shapes from only weak shape-level tags. It achieves this through a deep neural network trained simply to classify the shape as possessing or lacking the part. The novel structure of this network, which forms our core technical contribution, encourages finding large consistent regions across shapes that characterize the differentiating part. We also presented compelling results on strongly supervised segmentation using the same network.

There are several avenues for future work enabled by unstructured user annotations in public online 3D repositories. It would be interesting to leverage natural language processing in addition to geometric analysis to automatically infer salient shape tags and corresponding parts from free-form shape descriptions provided by people. It would also be interesting to generalize our network architecture to handle a larger and more heterogeneous sets of tags, and to scale robustly to multi-label settings.

\bibliographystyle{ieee}
\bibliography{ms}

\clearpage
\title{\Large
  {\bf Tags2Parts: Discovering Semantic Regions from Shape Tags} \\
  \vspace{3mm}{\em Supplementary Material}
}
\date{\vspace{-5mm}}
\author{}

\appendix

\maketitle

\section{Weakly-supervised segmentation precision-recall plots.}

\begin{figure}[b!]
  \vspace{-0.5cm}
  \centering
  \includegraphics[width=0.95\linewidth]{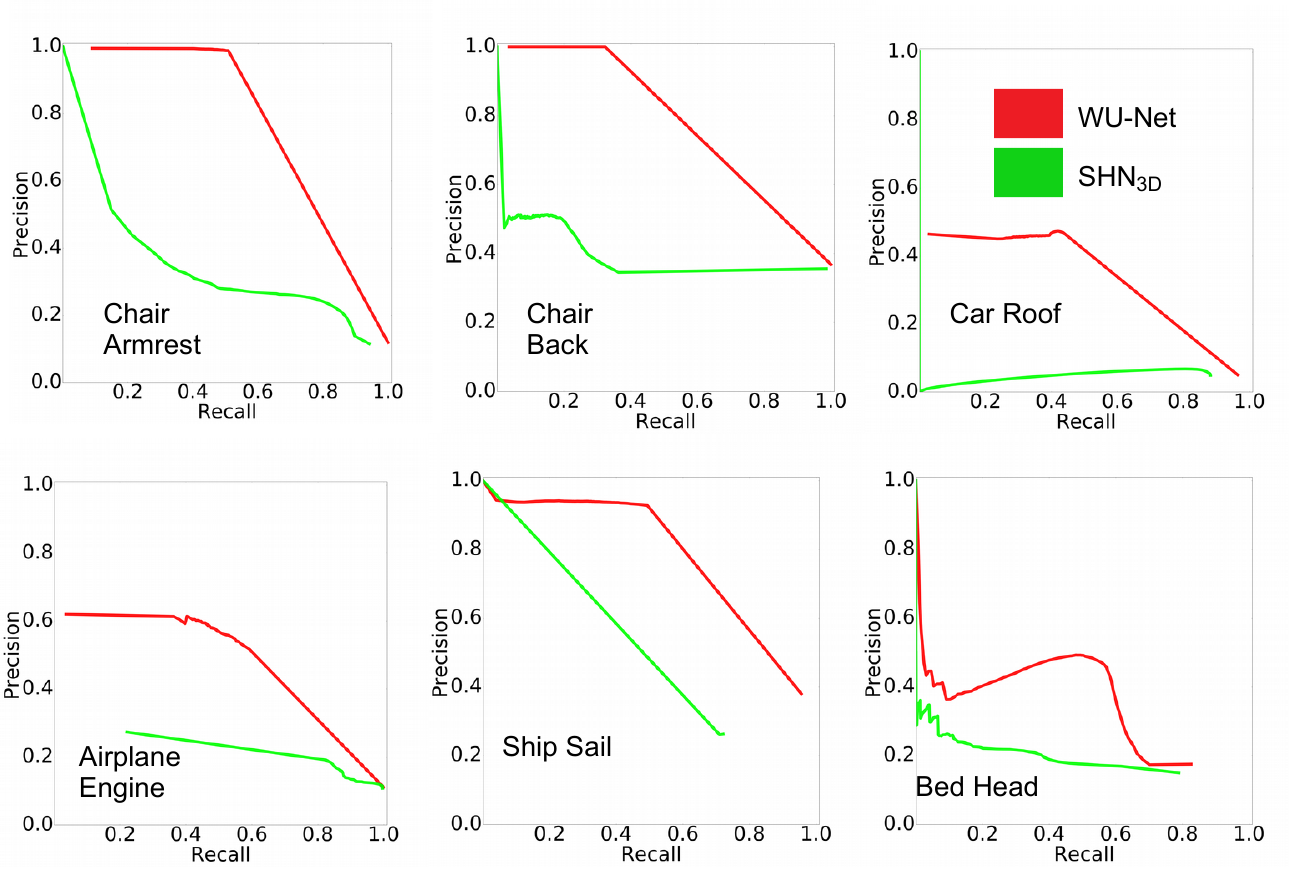}
  \caption{WU-Net (red) consistently outperforms a Stacked Hourglass Network SHN$_\text{3D}$ (3 deep U's without low resolution skip connections between different U's, green) on all categories (on training shapes, outputs symmetrized).}
  \label{fig:shn}
  \vspace{-0.5cm}
\end{figure}

\begin{figure*}[t!]
\vspace{-2mm}
\includegraphics[width=1.0\linewidth]{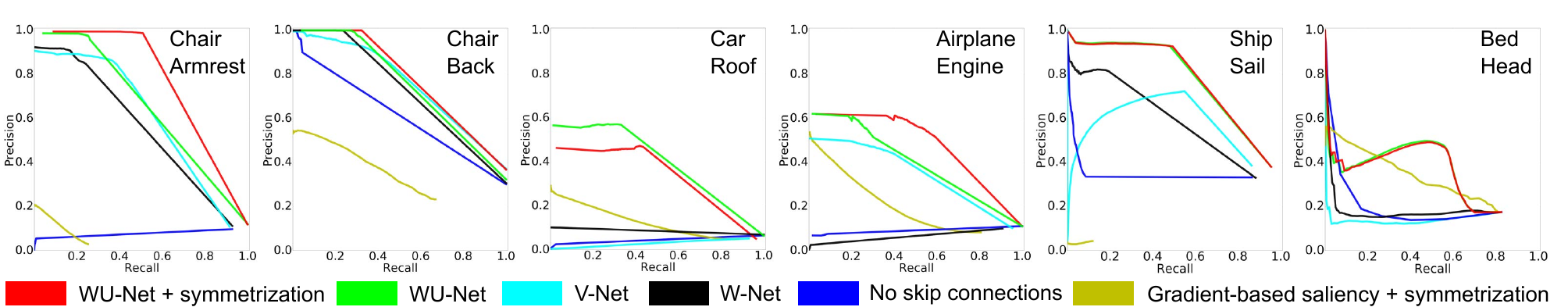}
\caption{\wunet vs various ablations for weakly-supervised segmentation (on training shapes).}
\label{fig:ablation_symmet}
\end{figure*}

\begin{figure*}[t!]
\vspace{-1mm}
\includegraphics[width=1.0\linewidth]{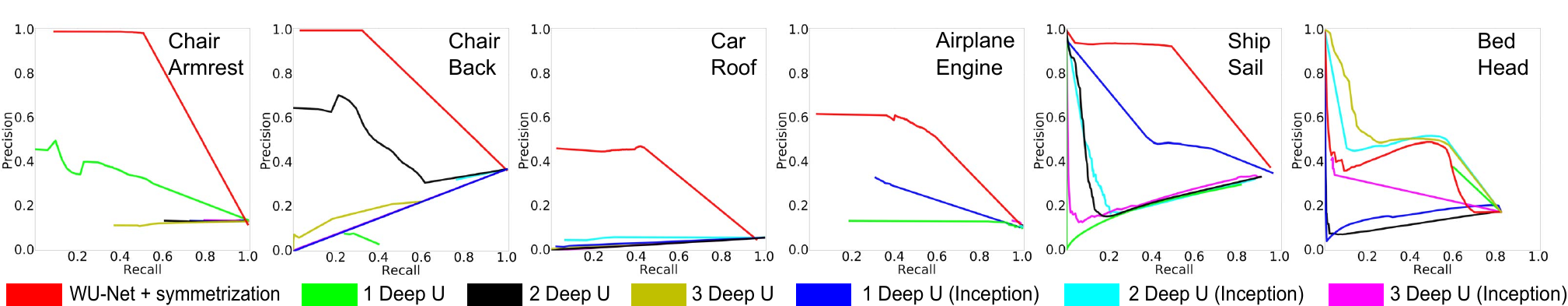}
\caption{\wunet vs Deep U alternatives (symmetrized) for weakly-supervised segmentation (on training shapes).}
\label{fig:deepnetworks}
\end{figure*}

\begin{figure*}[h!]
\vspace{-1mm}
\includegraphics[width=1.0\linewidth]{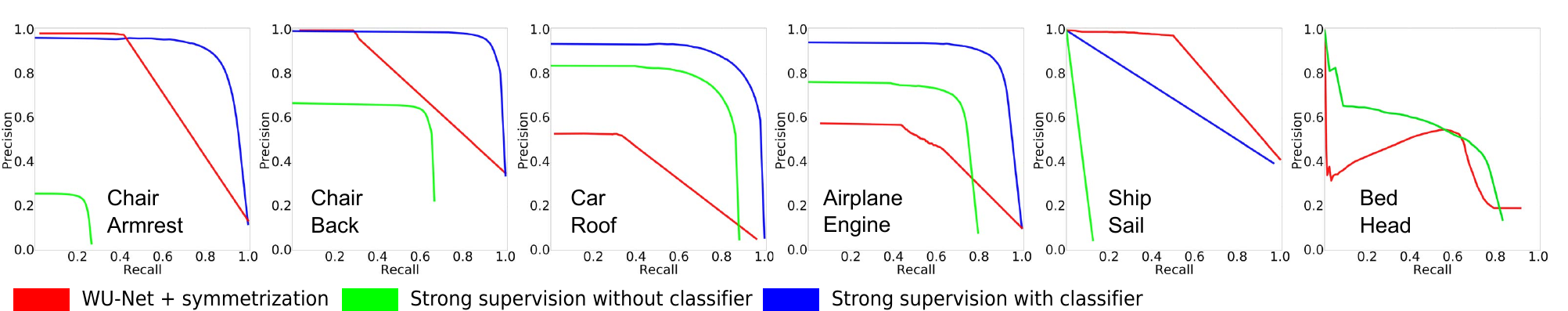}
\caption{Weakly supervised \wunet vs a strongly supervised baseline (on test shapes).}
\label{fig:strong}
\end{figure*}

\begin{figure*}[h!]
\vspace{-1mm}
\includegraphics[width=1.0\linewidth]{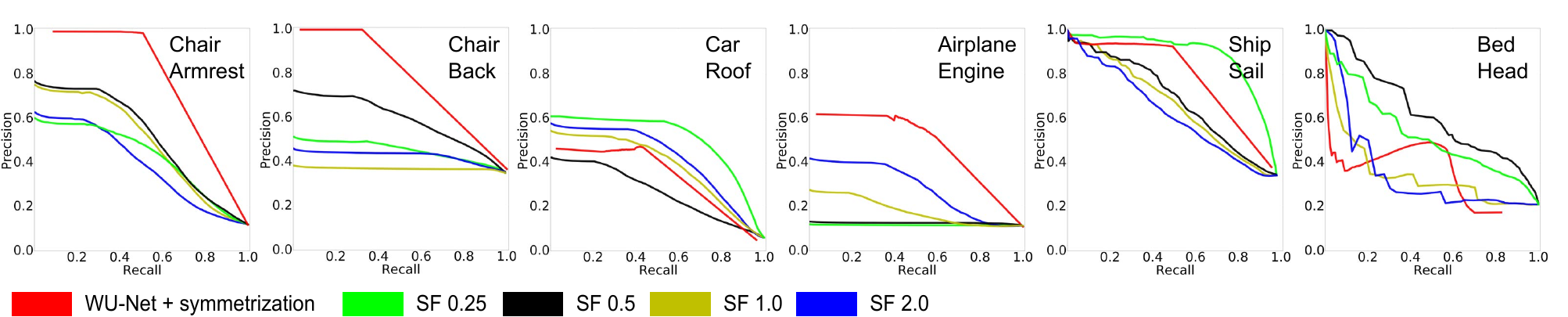}
\caption{\wunet vs Shilane and Funkhouser (SF)~\shortcite{Shilane2007} at different scales (on training shapes). Note that SF requires knowledge of ground truth tags at test time, whereas our method does not use them.}
\label{fig:shilane}
\end{figure*}

\begin{figure*}[h!]
\vspace{-1mm}
\includegraphics[width=1.0\linewidth]{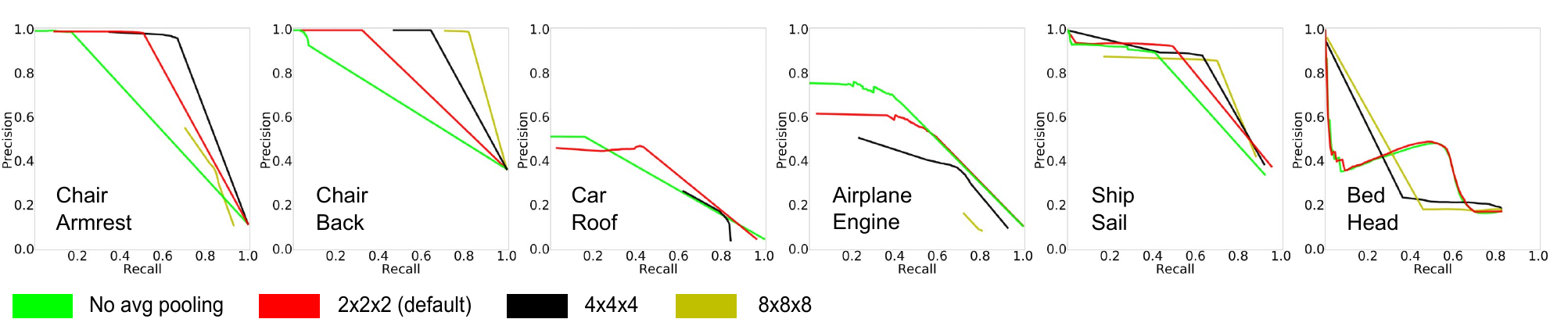}
\caption{The statistical effect of increasing the kernel size for average pooling at the end of \wunet.}
\label{fig:pooling}
\vspace{-5mm}
\end{figure*}

In Figures \ref{fig:shn}, \ref{fig:ablation_symmet}, \ref{fig:deepnetworks}, \ref{fig:strong}, \ref{fig:shilane}, \ref{fig:pooling}, we present full precision-recall plots from various experiments on weakly-supervised segmentation, including ablation studies and comparisons to prior work. The area under the curve (AUC) statistics summarizing these plots are presented in the main paper.

\section{More segmentation statistics and complete visualizations}

Tables \ref{tab:fullsup_cats}, \ref{tab:rotatedfullsup}, \ref{tab:newsplitfullsup} and \ref{tab:ioutable} show the per-category performance of different deep and shallow \wunet variants for strongly-supervised segmentation on the standard ShapeNet dataset, on (a) the train/test splits from Kalogerakis et al.~\cite{Kalogerakis17}, (b) on randomly rotated versions of the shapes in these splits, and (c) the splits from the recent ICCV challenge~\cite{Yi17b} (both accuracy and IOU statistics) respectively. Different variants of {\wunet}-style networks are given abbreviated names: 3SU is a sequence of {\bf 3} {\bf s}hallow {\bf U}'s (i.e. \wunet), 1DUI is {\bf 1} single {\bf d}eep {\bf U} ({\bf I}nception-style).

Our project website for this paper has visualizations of all segmentations of shapes in our datasets under both weak and strong supervision of \wunet.

The performance of \wunet on weakly-supervised segmentation of test set shapes mirrors that on the training set, as can be seen in Table 4 of the main paper as well as the visualizations on the website.

\clearpage

\clearpage

\begin{table*}[h!]
\small
 \centering
    \begin{tabular}{@{}c@{}||@{}c@{}|@{}c@{}||@{}c@{}|@{}c@{}|@{}c@{}|@{}c@{}|@{}c@{}|@{}c@{}|@{}c@{}|@{}c@{}|@{}c@{}|@{}c@{}|@{}c@{}|@{}c@{}}
      \,\,Category\,\, & \,\,\#train/\,\, & \,\,\#labels\,\, & \,\,Shape-\,\, & \,\,Shape-\,\, & \,\,1 SU\,\, & \,\,2 SU\,\, & \,\,3 SU\,\, & \,\,4 SU\,\, & \,\,1 DU\,\, & \,\,2 DU\,\, & \,\,3 DU\,\, & \,\,1 DUI\,\, & \,\,2 DUI\,\, & \,\,3 DUI\,\, \\
      ~ & \,\,\#test\,\, & ~ & \,\,Boost\,\, & \,\,PFCN\,\, & ~ & ~ & \,\,(WU-Net)\,\, & ~ & ~  & ~ & ~ & ~ & ~ & ~ \\
\hline
\hline
Airplane & 250/250 & 4 & 84.1 & 88.4 & 89.54 & 90.32 & 90.13 & 90.66 & 90.34 & 90.75 & 90.77 & 91.12 & 90.87 & {\bf{90.97}} \\
\hline
Bag & 38/38 & 2 & 94.3 & 95.5 & 93.24 & {\bf{96.51}} & 96.02 & 95.53 & 96.18 & 95.26 & 96.21 & 96.24 & 96.15 & 96.19 \\
\hline
Bike & 101/101 & 6 & 78.6 & {\bf{87.5}} & 80.84 & 84.07 & 84.77 & 85.75 & 85.04 & 85.04 & 85.54 & 85.19 & 83.66 & 85.9 \\
\hline
Cap & 27/28 & 2 & {\bf{94.8}} & 92 & 87.77 & 88.7 & 89.82 & 88.63 & 90.27 & 91.19 & 86.31 & 90.75 & 91.69 & 91.88 \\
\hline
Car & 250/250 & 4 & 75.5 & 86.6 & 88.91 & 89.61 & 89.44 & 89.67 & 89.7 & 89.87 & 90.02 & 90.13 & 90.15 & {\bf{90.17}} \\
\hline
Chair & 250/250 & 4 & 71.9 & 83.7 & 90.47 & 92.01 & 91.82 & 91.9 & 92.24 & 92.01 & 92.11 & 92.32 & 92.3 & {\bf{92.4}} \\
\hline
Earphone & 34/35 & 3 & 76 & {\bf{82.9}} & 67.21 & 75.86 & 78.53 & 74.44 & 78.24 & 80.84 & 74.71 & 82.2 & 77.34 & 79.73 \\
\hline
Guitar & 250/250 & 3 & 86.9 & 89.7 & 95.89 & 96.22 & 95.98 & 96.09 & 96.22 & 96.23 & 96.19 & 96.26 & 96.23 & {\bf{96.29}} \\
\hline
Knife & 196/196 & 2 & 84.1 & 87.1 & 83.81 & 90.33 & 90.96 & {\bf{92.42}} & 91.57 & 91.34 & 91.37 & 91.69 & 91.83 & 90.91 \\
\hline
Lamp & 250/250 & 4 & 63.8 & 78.3 & 75.75 & 77.97 & 77.37 & 80.91 & 82.7 & 83.63 & 82.96 & 84.38 & 83.82 & {\bf{85.09}} \\
\hline
Laptop & 222/222 & 2 & 79.4 & 95.2 & {\bf{96.86}} & 96.57 & 96.61 & 96.63 & 96.33 & 96.48 & 96.51 & 96.56 & 96.62 & 96.84 \\
\hline
Mug & 92/92 & 3 & 98.1 & 98.1 & 98.94 & 99.09 & 99.05 & {\bf{99.17}} & 99.14 & 98.81 & 99.16 & 99.16 & 99.14 & 99.15 \\
\hline
Pistol & 137/138 & 3 & 84.9 & 92.2 & 94.46 & 96.01 & 95.75 & 96.05 & 96.41 & 96.51 & {\bf{96.7}} & 96.55 & 96.54 & 96.55 \\
\hline
Rocket & 33/33 & 3 & {\bf{83.2}} & 81.5 & 75.64 & 75.35 & 79.94 & 75.36 & 76.29 & 76.98 & 75.61 & 77.93 & 78.05 & 79.73 \\
\hline
\,\,Skateboard\,\, & 76/76 & 3 & 89.6 & 92.5 & 94.54 & 94.32 & {\bf{94.66}} & 94.23 & 93.91 & 92.97 & 93.62 & 94.33 & 94.36 & 94.36 \\
\hline
Table & 250/250 & 3 & 83.9 & 92.5 & 90.33 & 93.58 & 92.91 & 92.99 & 93.94 & 93.32 & 94.37 & 94.57 & {\bf{94.92}} & 94.42 \\
\hline
\hline
      \multicolumn{3}{l|}{Category average} & 83.07 & 88.98 & 87.76 & 89.78 & 90.24 & 90.03 & 90.53 & 90.7 & 90.14 & 91.21 & 90.85 & {\bf{91.29}} \\
      \hline
  \end{tabular}
  \caption{Dataset statistics and strongly-supervised segmentation and labeling accuracy per category for test shapes in ShapeNetCore, versus ShapePFCN~\cite{Kalogerakis17} and ShapeBoost~\cite{Kalogerakis2010}, using the splits from \cite{Kalogerakis17}.}
  \label{tab:fullsup_cats}
	\vspace{3mm}
\end{table*}

\begin{table*}[h!]
\small
 \centering
    \begin{tabular}{@{}c@{}||@{}c@{}|@{}c@{}||@{}c@{}|@{}c@{}|@{}c@{}|@{}c@{}|@{}c@{}|@{}c@{}|@{}c@{}|@{}c@{}|@{}c@{}|@{}c@{}|@{}c@{}|@{}c@{}}
      \,\,Category\,\, & \,\,\#train/\,\, & \,\,\#labels\,\, & \,\,Shape-\,\, & \,\,Shape-\,\, & \,\,1 SU\,\, & \,\,2 SU\,\, & \,\,3 SU\,\, & \,\,4 SU\,\, & \,\,1 DU\,\, & \,\,2 DU\,\, & \,\,3 DU\,\, & \,\,1 DUI\,\, & \,\,2 DUI\,\, & \,\,3 DUI\,\, \\
      ~ & \,\,\#test\,\, & ~ & \,\,Boost\,\, & \,\,PFCN\,\, & ~ & ~ & \,\,(WU-Net)\,\, & ~ & ~  & ~ & ~ & ~ & ~ & ~ \\
\hline
\hline
Airplane & 250/250 & 4 & 84.1 & {\bf{88.4}} & 70.15 & 80.76 & 78.83 & 76.99 & 79.77 & 80.54 & 78.54 & 79.97 & 81.29 & 80.88 \\
\hline
Bag & 38/38 & 2 & 94.3 & {\bf{95.5}} & 93.67 & 93.2 & 92.45 & 93.49 & 93.09 & 93.5 & 93.01 & 93.04 & 93.69 & 93.22 \\
\hline
Bike & 101/101 & 6 & 78.6 & {\bf{87.5}} & 73.17 & 71.98 & 73.86 & 74.63 & 71.73 & 72.65 & 73.35 & 71.92 & 72.18 & 72.22 \\
\hline
Cap & 27/28 & 2 & {\bf{94.8}} & 92 & 73.43 & 70.79 & 72.98 & 72.54 & 70.23 & 68.22 & 74.31 & 73.37 & 72.84 & 75.03 \\
\hline
Car & 250/250 & 4 & 75.5 & {\bf{86.6}} & 74.66 & 76.12 & 78.03 & 78.3 & 78.42 & 80.02 & 80.25 & 77.58 & 78.8 & 78.49 \\
\hline
Chair & 250/250 & 4 & 71.9 & {\bf{83.7}} & 55.32 & 66.14 & 69.62 & 74.99 & 79.77 & 77.9 & 80.51 & 78.68 & 81.01 & 80.87 \\
\hline
Earphone & 34/35 & 3 & 76 & {\bf{82.9}} & 61.93 & 65.75 & 66.98 & 65.44 & 66.9 & 68.52 & 66.02 & 65.97 & 64.42 & 66.4 \\
\hline
Guitar & 250/250 & 3 & 86.9 & 89.7 & 88.54 & 91.91 & 92.16 & 93.01 & 93.06 & {\bf{94.17}} & 93.25 & 93.65 & 93.4 & 93.66 \\
\hline
Knife & 196/196 & 2 & {\bf{84.1}} & 87.1 & 71.33 & 71.55 & 70.24 & 71.01 & 79.24 & 80.25 & 78.04 & 80.05 & 79.02 & 79.61 \\
\hline
Lamp & 250/250 & 4 & 63.8 & {\bf{78.3}} & 58.5 & 58.65 & 60.63 & 64.05 & 66.68 & 65.87 & 68.64 & 71.14 & 69.34 & 70.98 \\
\hline
Laptop & 222/222 & 2 & 79.4 & {\bf{95.2}} & 53.91 & 56.36 & 54.39 & 51.64 & 57.12 & 50.6 & 57.71 & 62.18 & 62.64 & 62.43 \\
\hline
Mug & 92/92 & 3 & {\bf{98.1}} & {\bf{98.1}} & 95.77 & 97.29 & 97.5 & 97.58 & 96.86 & 97.26 & 96.14 & 96.55 & 96.41 & 96.37 \\
\hline
Pistol & 137/138 & 3 & 84.9 & {\bf{92.2}} & 67.2 & 61.88 & 66.69 & 65.06 & 77.14 & 77.87 & 76.92 & 74.59 & 74.71 & 74.11 \\
\hline
Rocket & 33/33 & 3 & {\bf{83.2}} & 81.5 & 71.96 & 70.85 & 69.26 & 70.72 & 67.12 & 69.24 & 68.43 & 67.59 & 72.09 & 69.86 \\
\hline
Skateboard & 76/76 & 3 & 89.6 & {\bf{92.5}} & 84.98 & 85.5 & 85.08 & 84.85 & 82.31 & 82.3 & 86.42 & 80.84 & 85.51 & 82.92 \\
\hline
Table & 250/250 & 3 & 83.9 & {\bf{92.5}} & 74.77 & 76.9 & 73.45 & 75.34 & 85.27 & 86.32 & 86.89 & 86.19 & 87.29 & 87.05 \\
\hline
\hline

      \multicolumn{3}{l|}{Category average} & 83.07 & {\bf{88.98}} & 73.08 & 74.73 & 75.14 & 75.6 & 77.79 & 77.83 & 78.65 & 78.33 & 79.04 & 79.01 \\
      \hline
  \end{tabular}
  \caption{Dataset statistics and strongly-supervised segmentation and labeling accuracy per category for randomly rotated test shapes in ShapeNetCore, versus ShapePFCN~\cite{Kalogerakis17} and ShapeBoost~\cite{Kalogerakis2010}, on the splits from \cite{Kalogerakis17}.}
  \label{tab:rotatedfullsup}
	\vspace{3mm}
\end{table*}

\begin{table*}[h!]
\small
 \centering
    \begin{tabular}{@{}c@{}||@{}c@{}|@{}c@{}||@{}c@{}|@{}c@{}|@{}c@{}|@{}c@{}|@{}c@{}|@{}c@{}|@{}c@{}|@{}c@{}|@{}c@{}|@{}c@{}}
      \,\,Category\,\, & \,\,\#train/\,\, & \,\,\#labels\,\, & \,\,1 SU\,\, & \,\,2 SU\,\, & \,\,3 SU\,\, & \,\,4 SU\,\, & \,\,1 DU\,\, & \,\,2 DU\,\, & \,\,3 DU\,\, & \,\,1 DUI\,\, & \,\,2 DUI\,\, & \,\,3 DUI\,\, \\
      ~ & \,\,\#test\,\, & ~ & ~ & ~ & \,\,(WU-Net)\,\, & ~ & ~ & ~ & ~ & ~ & ~ & ~ \\
\hline
\hline
Airplane & 1958/341 & 4 & 87.46 & 89.18 & 89.61 & 89.73 & 90.4 & 90.16 & 90.32 & 90.51 & 90.65 & {\bf{90.74}} \\
\hline
Bag & 54/14 & 2 & 93.51 & 90.96 & 93.44 & 92.83 & 96.02 & 95.57 & 95.74 & 96.12 & 96.43 & {\bf{96.56}} \\
\hline
Bike & 125/51 & 6 & 75.35 & 75.89 & 86.36 & 86.9 & 73.64 & 77.78 & {\bf{87.1}} & 86.95 & 86.1 & 86.83 \\
\hline
Cap & 39/11 & 2 & 88.32 & 87.01 & 87.4 & 87.24 & 83.35 & 86.72 & 88.38 & {\bf{90.62}} & 87.47 & 85.63 \\
\hline
Car & 659/158 & 4 & 89.24 & 90.34 & 90.41 & {\bf{90.49}} & 90.25 & 89.94 & 90.19 & 90.24 & 90.21 & 90.42 \\
\hline
Chair & 2658/704 & 4 & 91.16 & 92.93 & 93.13 & 93.4 & 93.85 & 93.91 & 93.92 & 93.95 & {\bf{94}} & 93.98 \\
\hline
Earphone & 49/14 & 3 & 70.54 & 90.35 & 91.6 & 91.61 & 91.68 & 92.1 & {\bf{92.52}} & 87.3 & 91.5 & 91.64 \\
\hline
Guitar & 550/159 & 3 & 95.63 & 95.75 & 95.65 & 95.89 & {\bf{96.11}} & 95.96 & 95.66 & 95.88 & 96.05 & 95.89 \\
\hline
Knife & 277/80 & 2 & 83.28 & 90.8 & 91.98 & 90.7 & 91.93 & 91.08 & 90.77 & {\bf{92.31}} & 91.83 & 91.49 \\
\hline
Lamp & 1118/296 & 4 & 73.87 & 78.21 & 78.38 & 80.49 & {\bf{88.27}} & 88.18 & 87.19 & 88.1 & 86.47 & 87.83 \\
\hline
Laptop & 324/83 & 2 & 96.66 & 96.88 & 96.79 & {\bf{97.23}} & 96.15 & 95.9 & 95.56 & 96.73 & 96.51 & 96.8 \\
\hline
Mug & 130/38 & 3 & 99.29 & 99.43 & 99.42 & 99.39 & 99.44 & {\bf{99.46}} & 99.4 & 99.34 & 99.38 & 99.43 \\
\hline
Pistol & 209/44 & 3 & 92.94 & 94.46 & 94.33 & 94.85 & 95.84 & 95.98 & {\bf{96.01}} & 95.91 & 95.85 & 96 \\
\hline
Rocket & 46/12 & 3 & {\bf{75.11}} & 73.67 & 74.46 & 74.76 & 72.93 & 69.85 & 74.94 & 71.45 & 74.01 & 74.2 \\
\hline
Skateboard & 106/31 & 3 & 94.5 & 94.13 & 93.73 & 94.1 & 94.68 & {\bf{94.85}} & 94.57 & 94.19 & 93.9 & 94.34 \\
\hline
Table & 3835/848 & 3 & 85.25 & 87.03 & 88.94 & 88.14 & 93 & 92.64 & 92.17 & {\bf{94.38}} & 93.78 & 92.22 \\
\hline
\hline

      \multicolumn{3}{l|}{Category average} & 87.01 & 89.19 & 90.35 & 90.48 & 90.47 & 90.63 & {\bf{91.53}} & 91.5 & 91.51 & 91.5 \\
      \hline
  \end{tabular}
  \caption{Dataset statistics and strongly-supervised segmentation and labeling accuracy per category for test shapes in ShapeNetCore on the new splits from the ShapeNet ICCV Challenge~\cite{Yi17b}.}
  \label{tab:newsplitfullsup}
	\vspace{3mm}
\end{table*}

\begin{table*}[h!]
\small
 \centering
    \begin{tabular}{@{}c@{}|@{}c@{}|@{}c@{}|@{}c@{}|@{}c@{}|@{}c@{}|@{}c@{}|@{}c@{}|@{}c@{}|@{}c@{}|@{}c@{}|@{}c@{}|@{}c@{}|@{}c@{}|@{}c@{}|@{}c@{}|@{}c@{}|@{}c@{}}
     Method & \,\,avg\,\, & \,\,plane\,\, & \,\,bag\,\, & \,\,cap\,\, & \,\,car\,\, &  \,\,chair\,\, & \,\,earphone\,\, & \,\,guitar\,\, & \,\,knife\,\, & \,\,lamp\,\, & \,\,laptop\,\, & \,\,bike\,\, & \,\,mug\,\, & \,\,pistol\,\, &  \,\,rocket\,\, & skateboard & \,\,table\,\, \\
\hline
\hline
2 DUI & 83.13 & 81 & 83.2 & 75.25 & 75.22 & 89.04 & 71.68 & 89.24 & 84.15 & 75.85 & 93.46 & 69.27 & 94.19 & 81.38 & 54.34 & 73.39 & 80.86 \\
\hline
1 DUI & 82.89 & 80.83 & 81.8 & 81.16 & 75.31 & 89.02 & 60.22 & 88.93 & 84.99 & 77.72 & 93.37 & 70.12 & 93.9 & 82 & 50.28 & 72.46 & 80.12 \\
\hline
3 DUI & 81.12 & 80.9 & 83.4 & 73.1 & 75.93 & 89.04 & 73.94 & 89.13 & 83.59 & 77.1 & 93.51 & 69.44 & 94.7 & 82.61 & 47.32 & 72.7 & 74.52 \\
\hline
3 DU & 81.13 & 80.35 & 78.91 & 76.67 & 75.63 & 88.87 & 74.96 & 88.16 & 82.94 & 75.56 & 92.53 & 69.17 & 94.49 & 82.22 & 49.6 & 73.77 & 75.39 \\
\hline
1 DU & 81.03 & 80.42 & 81.75 & 68.82 & 73.95 & 88.81 & 73.97 & 89.82 & 84.34 & 76.64 & 92.85 & 29.7 & 94.78 & 82.18 & 48.69 & 71.95 & 77.47 \\
\hline
2 DU & 80.97 & 79.91 & 79.56 & 74 & 74.36 & 88.75 & 70.5 & 89.08 & 83.12 & 75.45 & 92.93 & 34.75 & 94.92 & 82.95 & 46.14 & 74.15 & 77.38 \\
\hline
4 SU & 79.84 & 77.19 & 68.79 & 76.52 & 75.01 & 87.41 & 68.59 & 89.54 & 83.16 & 68.08 & 93.96 & 68.63 & 94.28 & 80.07 & 50.07 & 71.57 & 74.91 \\
\hline
3 SU & 79.35 & 76.51 & 71.2 & 76.39 & 75.07 & 86.95 & 69.24 & 89.08 & 84.64 & 66.87 & 93.26 & 61.91 & 94.66 & 77.18 & 51.56 & 71.04 & 74.62 \\
\hline
2 SU & 77.87 & 75.31 & 59.01 & 74.68 & 74.35 & 86.24 & 68.46 & 89.15 & 82.75 & 63.03 & 93.34 & 25.04 & 94.71 & 77.49 & 44.69 & 71.73 & 74.32 \\
\hline
1 SU & 74.35 & 70.86 & 73.46 & 78.06 & 69.05 & 81.31 & 37.21 & 88.49 & 71.97 & 58.67 & 93.01 & 24.66 & 93.73 & 74.71 & 46.46 & 73.23 & 72.23 \\
\hline
\hline
  \end{tabular}
  \caption{IOU scores for different versions of Deep and Shallow-U networks for strongly-supervised segmentation of test shapes in ShapeNetCore on the new splits from the ShapeNet ICCV Challenge~\cite{Yi17b}.}
  \label{tab:ioutable}
	\vspace{3mm}
\end{table*}

\end{document}